\documentclass[10pt,letterpaper]{article}

\usepackage{cogsci}
\usepackage{pslatex}
\usepackage{apacite}
\usepackage{graphicx}
\usepackage{makecell}
\usepackage{multicol}
\usepackage{multirow}
\usepackage{longtable}
\LTcapwidth=\textwidth
\usepackage{caption}
\usepackage{subcaption}
\usepackage{amsmath}
\usepackage{xcolor}
\usepackage{outlines}
\title{On the Cognition of Visual Question Answering Models and Human Intelligence: A Comparative Study}

\author{\large \bf Liben Chen(lc4438@nyu.edu) \\ \large \bf Long Chen (lc3424@nyu.edu) \\ \large \bf Tian Ellison-Chen (te2049@nyu.edu) \\ \large \bf Zhuoyuan Xu (zx1137@nyu.edu) \\Center for Data Science, New York University, New York, NY 10012 USA}

\begin{document}

\maketitle

\begin{abstract}
Visual Question Answering (VQA) is a challenging task that requires cross-modal understanding and reasoning of visual image and natural language question. To inspect the association of VQA models to human cognition, we designed a survey to record human thinking process and analyzed VQA models by comparing the outputs and attention maps with those of humans. We found that although the VQA models resemble human cognition in architecture and performs similarly with human on the recognition-level, they still struggle with cognitive inferences. The analysis of human thinking procedure serves to direct future research and introduce more cognitive capacity into modeling features and architectures.  

\end{abstract}

\section{Introduction}
The task of Visual Question Answering (VQA) leverages both the visual and the textual analysis to answer a free-form, open-ended, natural-language question with an image~\cite{antol2015vqa}. The fields of computer vision and natural language processing have attained much success to reach almost human-level performance separately, such as YOLOR in object recognition ~\cite{wang2021you} and RoBerta on the GLUE benchmark ~\cite{liu2019roberta}. Still, combining them together to solve multi-discipline problems such as VQA is yet a challenging task. Furthermore, such a combination may not only cost much more computational power but also require the development of new features and model architectures. A VQA model needs to extract the visual and textual representations as well as their cross-modal interactions to inform the final answer. Most importantly, the VQA task might need commonsense reasoning and semantic knowledge that are natural to human cognitive abilities. A simple question for humans to answer with their specific cognitive abilities might be difficult for the machine.

In pursuit of the cognitive plausibility, we are interested in the association of VQA models to actual human cognition and perception from three perspectives: (1) model architecture inspired by the human thought process, (2) representation and collaborative understanding of visual and textual information, and (3) prediction results and their rationales. Based on our hypothesis of the human question answering process, we train a VQA baseline model and test a state-of-the-art (SOTA) model to predict answers for a given image and question pair, and develop a survey to collect human thought processes on a list of representative questions. By comparing outputs and attention maps from the baseline and SOTA model to those from humans, we are able to assess whether the deep learning approaches are robust resemblances of human results and possibly human thinking processes to identify targets and answer questions, though we expect them to fail to capture some semantic information and rationales that are key to human reasoning. Our findings reveal some similarities and differences between human reasoning and the VQA models. Moreover, they enlightened us with new research directions to augment deep learning model designs by creating new model architectures and training processes more intuitive to human cognition and developing thoughts in performing more complicated, multi-modal tasks.

\section{Related Work}
\subsection{VQA and Datasets}
Recent studies on Visual Question Answering (VQA) have an emergent need for an accurate and comprehensive dataset. Most VQA datasets are built upon Microsoft Common Objects in Context (MS COCO), a widely used vision-textual dataset curated for object recognition in the context of scene understanding with abundant varieties~\cite{lin2014microsoft}. VQA 1.0 is a widely used dataset, which consists of the visual part of real images from MS COCO and abstract, animated scenes, as well as questions (often several for one image) and answers from human annotators~\cite{antol2015vqa}. It contains 22 types of questions, which covers most general cases of real-life visual question answering scenarios. There is also a VQA v2.0 dataset~\cite{goyal2017making}, which improves upon the original VQA 1.0 dataset with its correction of imbalanced answers, where the number of answers to a question is often skewed. The balancing procedure is that for an image-question-answer triplet ($I, Q, A$), human annotators are asked to identify a different but similar image $I'$ such that the answer $A'$ to the same question $Q$ is different. With such an annotation system, the dataset will have a more even distribution of answers for each type of question.
\subsection{VQA Model Architectures}
The recent development of Visual Question Answering models is built upon the maturity of visual and textual embedding models. The high-level concept of the VQA system is to integrate the visual and textual representations of inputs. Generally, there are four directions of development, as we will discuss here.

\vspace{-.3cm}
\paragraph{Joint Embedding} Joint embedding creates embeddings for images and questions and jointly trains them into a unified visual-textual representation. Most models utilize pre-trained visual models such as VGG-Net~\cite{simonyan2014very} and ResNet~\cite{he2016deep}. These models are trained to have a universal visual representation that suits the need of general VQA problems~\cite{hu2017learning, ma2015multimodal}. More recent studies have incorporated transformer architectures for image and text embeddings~\cite{li2020oscar, wang2021vlmo}. A variety of language models are used for textual representation. \citeA{ren2015faster} proposed the R-CNN architecture which uses VGG-Net and LSTM for object detection and downstream VQA tasks. \citeA{garderes2020conceptbert} proposed ConceptBert which utilizes BERT for textual embedding. To combine the two embeddings, multi-layer perceptrons are generally used. For transformer-based models, the transformer architecture will combine image and text into a unified representation.

\vspace{-.3cm}
\paragraph{Attention mechanism} Joint embedding is limited such that image representation is global without a more fine-grained and relevant focus on the text. The aim of the attention mechanism is to find local features in the context of questions to further enhance a coherent representation of the two embeddings. A simple attention mechanism with the LSTM model is used by \citeA{zhu2016visual7w}. Further, \cite{xu2016ask} proposed a multi-hop image attention scheme by employing attention on both word and question levels.

\vspace{-.3cm}
\paragraph{Compositional structure} Considering the two aforementioned approaches where monolithic representations for visual and textual inputs are modeled respectively, this approach, motivated by the categorical nature of questions in the VQA paradigm, aggregates models with a systematic selection of a sub-model by the context of model outputs. In this sense, the model is designed to have sub-models that can be fine-tuned to better serve the specific domain of one type of question, and the model will automatically select a sub-model for final output. \citeA{noh2016image} proposed a model with dynamic memory networks (DMN) to perform multiple passes, then using a joint loss over each of the passes.

\vspace{-.3cm}
\paragraph{Knowledge base} Intuitively, human perception involves abundant and pre-learned ``common sense'' and factual knowledge and uses such knowledge for specific tasks. The previously discussed approaches can only learn knowledge from the training set, which cannot achieve coverage for all real-life cases. This notion of transfer learning should conceptually help the model better analyze the inputs. Recent developments of large-scale knowledge bases, such as ConceptNet~\cite{speer2017conceptnet} and DBpedia~\cite{auer2007dbpedia}, promote the inclusion of such systems in VQA architectures. \citeA{wu2016ask} incorporated DBpedia with a joint embedding approach by retrieving and embedding external knowledge related to the textual and vision features with Doc2Vec, then feeding the knowledge-fused embedding into an LSTM model for interpretation.


\section{Method}
\subsection{Hypothesis}
Many model architectures for the VQA task are motivated by human cognitive abilities to comprehend a question associated with an image and reasoning answers. Cross-modal comprehension requires integration between recognition and cognition: after detecting targets and attributes from recognition, cognitive reasoning is needed to infer answers from observations. The modeling approach to simulate human cognition was inspired by~\citeA{pratl2005artificial} in building bionic models for an automation scenario recognition system. The authors relied on the notion that although the human brain memorizes images and scenarios throughout life, it is unable to analyze all the incoming stimuli in real-time. Instead, it perceives and recognizes characteristic forms of the external world backed by past memories. In regard to technically implementing the observed principle, the authors admitted that it would be impossible to construct an artificial system that implements every functional aspect of the architecture of the human brain. However, simpler solutions are feasible if constraints on the definition of brain capabilities are imposed such that a system only needs to understand images similar to already memorized ones and is capable in a limited set of scenarios and applications. Such ideology buttressed the cognitive relevance between the deep learning approach and the psychological mechanism, resembling the relationship between the training and test dataset, extracting feature embeddings from image and text, and specified tasks and output of the VQA model.

In our hypothesis, we presume that the human thought process includes but is not limited to: (1) having a general understanding of the image on the recognition level, (2) comprehending and allocating the subject of the question in the visual data on the recognition level and (3) reasoning the answer in the context of the question with the aid of image and experience on the cognitive level. A deep learning approach could be a robust simulation of the aforementioned thought process under an appropriate design. Its learned representation and output may be comparable to human behavioral data for assessment. It is able to perform the human cognitive task to a satisfactory accuracy. However, the approach is not necessarily meant to re-create full human intelligence. 

\begin{figure*}[ht]
\begin{center}
\includegraphics[width=18cm]{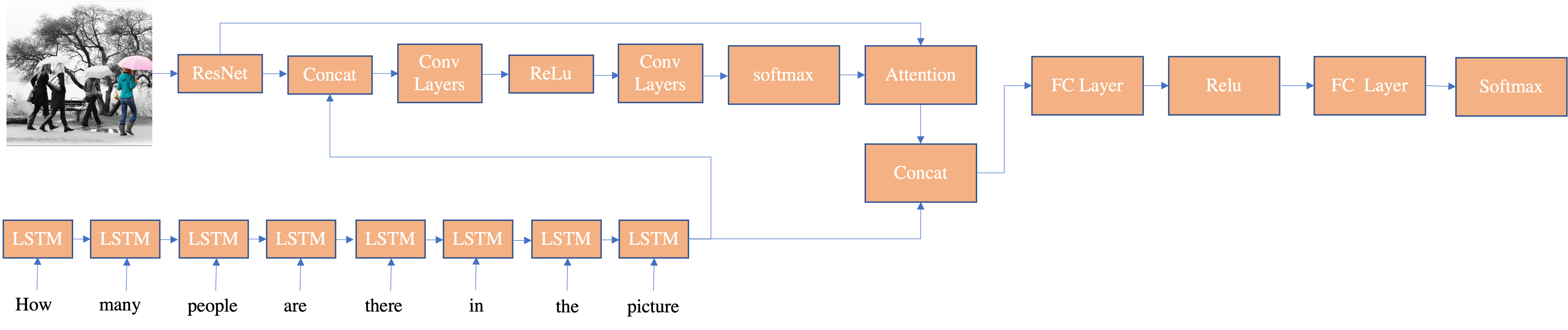}
\end{center}
\vspace{-.4cm}
\caption{Baseline model architecture.}
\label{sample-figure}
\end{figure*}

\subsection{Baseline Model}
\subsubsection{General Description} In this section, we introduce a strong baseline deep learning model on VQA \cite{kazemi2017show}. The model applies LSTM \cite{hochreiter1997long} with learnable weights to extract the textual features from the questions and prompts and the visual features from images by using pre-trained ResNet \cite{he2016deep}. Then, both visual and textual features are concatenated and piped into the convolutional neural network \cite{lecun1995convolutional} for further processing. An attention \cite{liu2018visual} module follows ResNet blocks to help the model better learn important parts of visual and textual features. The model is trained on the standard dataset VQA 1.0 and achieves adequate performance in evaluation on both VQA 1.0 and VQA 2.0 datasets. \citeA{kazemi2017show} proposed various baseline models in their papers through parameter tuning and different pretraining methods and we select the one with the strongest which achieves $61\%$ accuracy on the evaluation set from VQA 2.0 dataset. We start our analysis on this baseline model because the model has a relatively simple, primitive architecture to be investigated, while also being equipped with an interesting attention mechanism that has a potential counterpart in human perception.

\subsubsection{Model Specification} In this baseline model, we employ a uni-directional LSTM and use the context vector from the LSTM as the representation of the whole question. The context vector is then concatenated plainly with the feature matrix from the ResNet. The concatenation is done in a way where the dimension of the context vector is aligned with the number of columns of the feature matrix and the context vector is appended directly. After the concatenation, we have the attention module where we employ a $1 \times 1 \times 512$ convolutional layer on the feature matrix and a ReLU layer follows for providing non-linearity. In the attention module, we later scale down the dimension of the feature matrix by feeding the output from the ReLU layer to a $1 \times 1 \times 2$ convolutional layer. Then, the scaled-down feature matrix is treated with the softmax to output the attention weight matrix. We will describe the attention with more  mathematical details below. After the attention module, we have a fully-connected layer of dimension $1024$, followed by a ReLU layer. Finally, we have a fully connected layer of dimension $3000$ and use a softmax layer to model the probability distribution on output classes. We use the negative log-likelihood loss to train the whole architecture 
\begin{equation}
\ell = \frac{1}{K} \sum_{k=1}^{K} -\log P(a_k \vert I, q) 
\end{equation}
where $K$ is the total number of answers and $a_k$ is the $k^{th}$ answer. $I$ stands for features from images and $q$ represents the features from the question. 

\subsubsection{Attention Module} Attention module is an important part of the model worth investigating because we can potentially find counterpart mechanisms in human perception and perform comparisons between them. Thus, in this section, we provide mathematical details on the attention module. Specifically, given a feature matrix $\phi$, the element of the attended feature matrix $X_c$, is 
\begin{align}
    x_c &= \sum_{l} \alpha_{c,l} \phi_{l}\\
    s.t&. \sum^{L}_{l=1} \alpha_{l,c} = 1
\end{align}

where $l$ is all spatial locations $l=\{1, \dots, L\}$ in the feature matrix $\phi$ and $\alpha_{l,c}$ is the attention weight at each spatial location. It is worth noting that $c$ is the index of the attention head, meaning that the number of attended feature matrix is a parameter can be tuned. With multiple attended feature matrix, we finally stack them to form the final attended matrix for later layers. Notice that the attention weight is mathematically calculated as
\begin{equation}
\alpha_{l,c} \propto \exp{F_{c}(s, \phi_{l})}
\end{equation}
where $F_c$ is the feature map from the two convolutional layers described in the model specification section above. 

\subsubsection{Dataset and Training Details} We used the standard VQA 1.0 dataset which has a predefined train, test, and validation splits. There are 204,721 images in the VQA 1.0 and all of them are from MS COCO dataset \cite{lin2014microsoft}. We applied the Adam optimizer \cite{kingma2014adam} with a batch size of 128. The learning rate was decayed according to $l_{step} = 0.5^{\frac{step}{decay\_steps}}l_{0}$, where we set $l_0$ as 0.001. We evaluated the result on the VQA 1.0 validation dataset and obtained a 60.5 accuracy score at the end. We consider this result is marginally different from the one reported in the original paper and thus the training is considered successful. 


\begin{figure*}
    \centering
    \includegraphics[width=.7\textwidth]{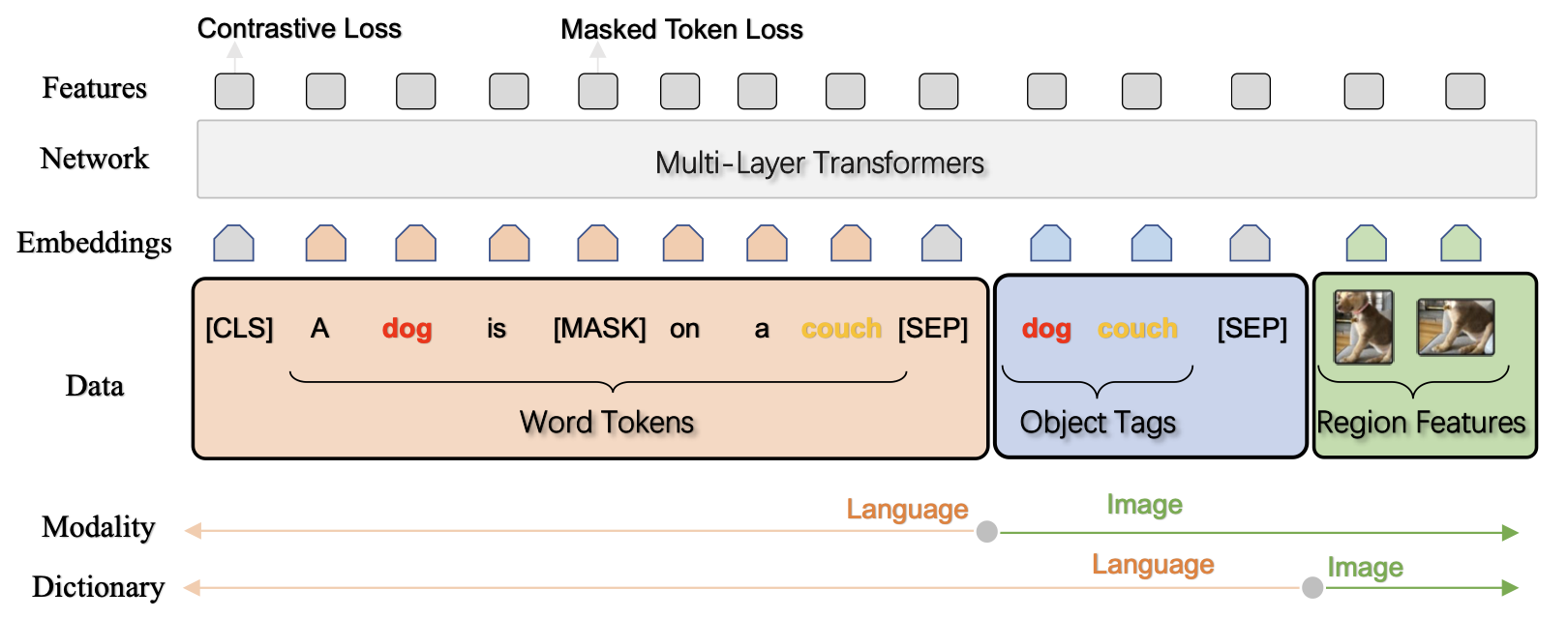}
    \vspace{-.3cm}
    \caption{Model architecture of OSCAR.}
    \label{fig:oscar_architecture}
\end{figure*}

\subsection{OSCAR}
OSCAR~\cite{li2020oscar}, which stands for Object-Semantics Aligned Pre-training for Vision-Language Tasks, is a recent development of VQA architecture with a transformer model. It uses object tags detected in images as anchor points to promote the learning of alignments. It includes semantics space, an embedding that maps an input to a vector that represents the semantics. This transformer architecture incorporates attention mechanism among textual embedding, image embedding and semantic embedding. With such a technique, the textual and visual features will be aligned by semantic similarity. It also utilizes region features from images as another layer of information. The model architecture is shown in Figure.~\ref{fig:oscar_architecture}. In this study, we utilize a pre-trained and fine-tuned OSCAR model for VQA purposes as a SOTA model to analyze its performance with our baseline model and with human perceptions. We aim to see some improvements in the robustness and accuracy of responses in comparison to the baseline.

\begin{figure*}[ht]
\begin{center}
\includegraphics[width=\linewidth]{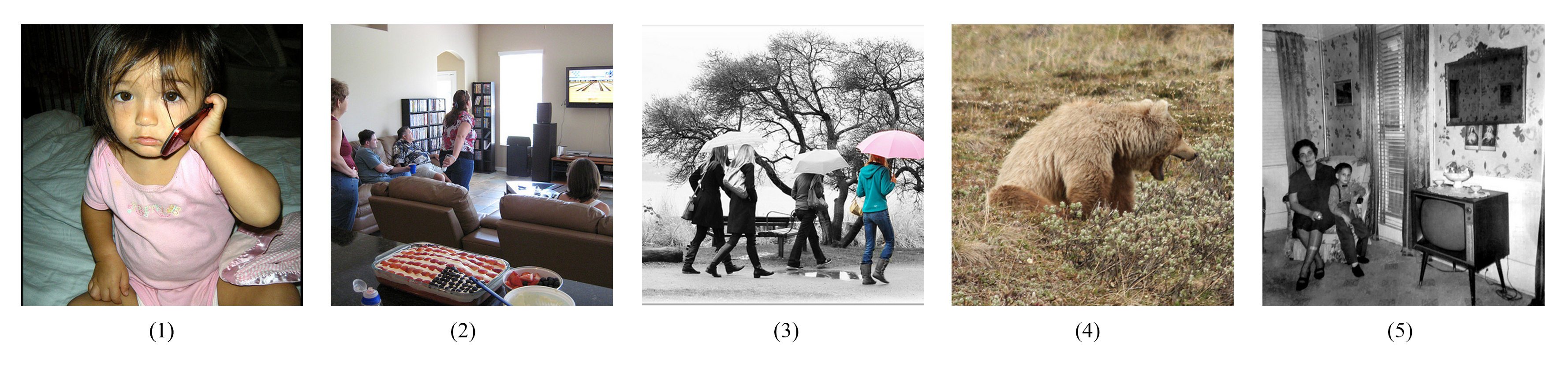}
\end{center}
\vspace{-.7cm}
\caption{5 survey images selected from MS COCO dataset.} 
\label{surveyimg}
\end{figure*}

\subsection{Survey Design}
We created a survey to capture the human thought process in VQA tasks, echoing our hypotheses and model architectures. The survey contains 5 images as shown in Fig.\ref{surveyimg} that we deem to cover a wide variety of the number of objects, image focus, color and texture,.etc. We asked 4 questions for each image, and the total 20 questions span the four categories in Table.\ref{tab:qtype} that testify to the major cognitive abilities of recognition and inference. The 4 question types are simplified from the studies in~\citeA{manmadhan2020visual} and~\citeA{shi2018question}. The list of questions is shown in Table.\ref{vqaq}. We limited the questions to be between 5 to 8 words to keep them straightforward, as prompt engineering is proven to affect the model performance~\cite{liu2021pre, jin2021good}. Our images and questions are close resemblances to those that have appeared in the official VQA dataset and have been widely used to train models in other paper~\cite{wang2021vlmo, yu2022coca}. Moreover, for each image-question pair, we let the participants answer a list of sub-questions in Table.\ref{embededq} to retrospect their reasoning process. The survey participants are students from the New York University Center of Data Science, or people who have studied machine learning and deep learning models in their degree work.

One powerful tool that helps us to understand the reasoning behind our prediction is the learned representation on the attention map in each step in the model with an attention mechanism. We first assume the attention of our models in this paper and that humans possess the same ability to uncover the distributions of importance level the model and human put on each patch of the test image. Then, we designed sub-questions 1 and 3 to determine a human attention map to compare with that from our model. Sub-question 1 focuses on human visual attention before viewing the natural-language VQA question, while sub-question 3 is asked after the question (as in sub-question 2) to observe the attention for any potential shift.

We use sub-question 4 to lead participants to think back on their rationale and any other information relevant to their responses. We hope to understand how people come to conclusions in the first 3 sub-questions and check if any inference they make is beyond the bare pixels and words. Meanwhile, we are curious about whether our participants may find the questions confusing to answer due to vagueness in-text wording and visual elements, bias from their common sense and knowledge and question types,.etc. Thus, we propose sub-questions 5 and 6 while we assume the time, or hesitation and contemplation in practice, appearing during the question answering process reflects the difficulty of the question to humans. With these investigative questions, we aim to create a thorough picture of human thought process in VQA tasks to compare with our machine. 

\begin{table*}[!ht]
\begin{center} 
\vskip 0.12in
\begin{tabular}{lll} 
\hline
Type    &  CV task & Example VQA question \\
\hline
Object-oriented        &   Object recognition & What is in the image? \\
 & Object detection & Are there any dogs in the image?\\
 & Attribute classification & What color is the umbrella?\\
Scene-oriented   &   Scene detection & Is it raining?\\
& Spatial relationship & What is between the cat and the sofa?\\
Counting & Counting & How many people are in the image?\\
Knowledge-oriented & Commonsense Inference & Does this person have 20/20 vision?\\
& Knowledge-based reasoning & Is this a vegetarian pizza?\\
\hline
\end{tabular} 
\end{center}
\vspace{-.3cm}
\caption{Question types.} 
\label{tab:qtype}
\end{table*}

\begin{table}[!ht]
\begin{center} 
\vskip 0.12in
\begin{tabular}{ll} 
\hline
Image    &  Question\\
\hline
1        &   (1) What is in the image? \\
 & (2) Is this baby crying?\\
 & (3) What is this baby doing?\\
 & (4) Is this baby a boy or girl?\\
2        &   (1) How many people are standing? \\
 & (2) What are people watching?\\
 & (3) What food is on the table?\\
 & (4) Where is the TV?\\
3        &   (1) Is it raining? \\
 & (2) How many people are there?\\
 & (3) What color are the umbrellas on the left?\\
 & (4) Are people in the forest?\\
4        &   (1) What animal is in the image?\\
 & (2) Is the bear hungry?\\
 & (3) What color is the bear's fur?\\
 & (4) Is the bear in the river?\\
5        &   (1) Who is older? \\
 & (2) Which room are they in?\\
 & (3) What is next to people?\\
 & (4) How many people are there?\\
\hline
\end{tabular} 
\end{center} 
\vspace{-.3cm}
\caption{VQA questions in the survey.}
\label{vqaq} 
\end{table}

\begin{table*}[!ht]
\begin{center} 
\vskip 0.12in
\begin{tabular}{ll} 
\hline
Number    &  Question\\
\hline
1        &   Please select the section in the image that attracts your attention at the first sight. \\
2 & Please answer the following question. \textit{Note: This points to the VQA question.}\\
3 & Please select the section in the image that contains the most information to answer the previous question.\\
4 & Please state in 1-3 sentences how you answer the previously given question. (e.g. What do you see in the \\
& image? Why do you focus on the sections you just selected? What induction/ educational guess have you made?\\
&  What else have you seen that is useful to answer the question?)\\
5        &  Have you hesitated when you answer the question based on the given image? (5: yes, for a long time; 1: no, \\
& immediately)\\
6 & If you took some time/thoughts to answer the question, what's the reason?\\
\hline
\end{tabular} 
\end{center}
\vspace{-.3cm}
\caption{Sub-questions under each image-question pair in the survey.}
\label{embededq} 
\end{table*}

\section{Results and Discussion}
\subsection{Human Responses}
We invited 21 participants to fill in our survey. We summarize their answers to each image-question pair together with those from the baseline model is in Table.\ref{tab:survey_result}. Since our VQA questions are mostly open-ended and our participants have different ways of framing their descriptions, we first extract keywords and phrases that are most relevant to each question from the answers. We assume human's writing style does not affect their decision-making and is beyond the topic of this paper. Then, we calculate the proportion of the number of times each keyword and phrase appear over the total number of answers we obtain for each question. From this table, we observe some general trends underlying human responses.

In most questions, a majority of answers share similar or closely related keywords. Yet, they include additional information people obtain via their cognitive abilities on elements in the image and background knowledge, introducing variations to the answers. For example, in Image 1 Question 1, all of the participants have the word "girl" in their answers, which itself is sufficient to answer the question "What is in the image?" Still, about $62\%$ responses include "phone" held on the girl's hand, and $24\%$ include the environment that the girl is "in the dark". In Image 5 Question 1, due to language habits, almost all participants indicated the gender of the person they consider older to be "female", but chose different wordings preferences like "woman", "lady", "female", "girl" and "mother" in their responses. Inferring from knowledge like in Image 1 Question 3, $76\%$ people believe the baby is "talking" or "answering" the phone while the image itself does not explicitly determine the girl's active usage of the phone. The action of talking is inferred from people's commonsense about what often happens when people are with a phone. We will further study these variations when we discuss attention and human decision-making process in the later section.

Despite keyword similarity in many questions, participants give relatively diverse responses in questions that have multiple and equally correct answers under human judgment. First, the question may point to multiple items in the image. In Image 2 Question 3, we see a wide variety of keywords including "fruit", "pie", "dessert", "pasta", "pizza" and "not sure", because (1) we have more than one food on the table and (2) people know various names for each food from experience. Second, the image does not provide enough information for a concrete answer and may be interpreted differently by human inference. As most portions of Image 3 are adjusted into grayscale, when we ask about the color of the umbrella on the left, participants' opinions spread across "white", "gray" and "not sure". In these cases, no answers can be considered a definite "false" from our perspective. Such vagueness may be natural to humans, but challenging to machine design.

Overall, nonetheless, the majority of participants responded with similar keywords or not, human answers would vary as long as the questions are open-ended and the answer is expected to be free-formed. They may share mutual information that uncovers underlying agreements among participants, e.g. the aforementioned example about the gender of the older person in Image 5 Question 1. Or, they may be completely different from the answers to the food on the table in Image 2 Question 3. This happens regardless of question types in Table.\ref{tab:qtype}. On the other hand, the VQA models only produce one succinct answer at a time. So, is it possible for them to perform well on human-like responses? Do they predict using a similar learning scenario with humans? If not, is it necessary to equip the human way of thinking in their learning? These observations lead us to expand our next stage of analysis into 3 steps: (1) direct comparison between human and model answers on our image-question pairs, (2) human and model attention, and (3) human decision-making process. We will summarize and provide evidence including cognitive theories and experiments on different stages in human reasoning via visual and textual stimulus. 

\subsection{Human Answer versus Model Prediction}
\begin{figure}[ht]
    \begin{subfigure}{\linewidth}
        \includegraphics[width=\textwidth]{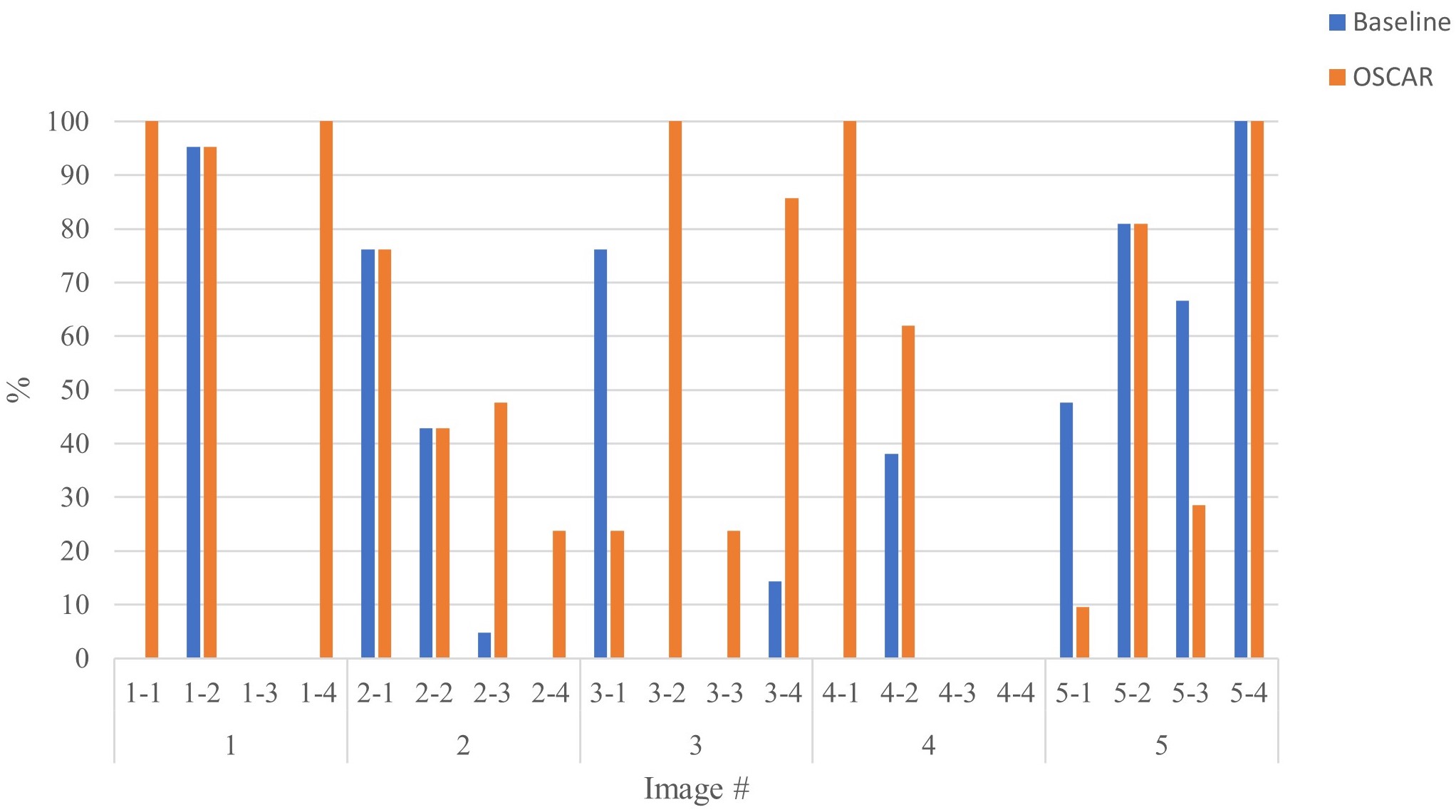}
        \vspace{-.5cm}
        \caption{Grouped by image number.}
        \label{imgnumpercent}
    \end{subfigure}
    \begin{subfigure}{\linewidth}
        \includegraphics[width=\textwidth]{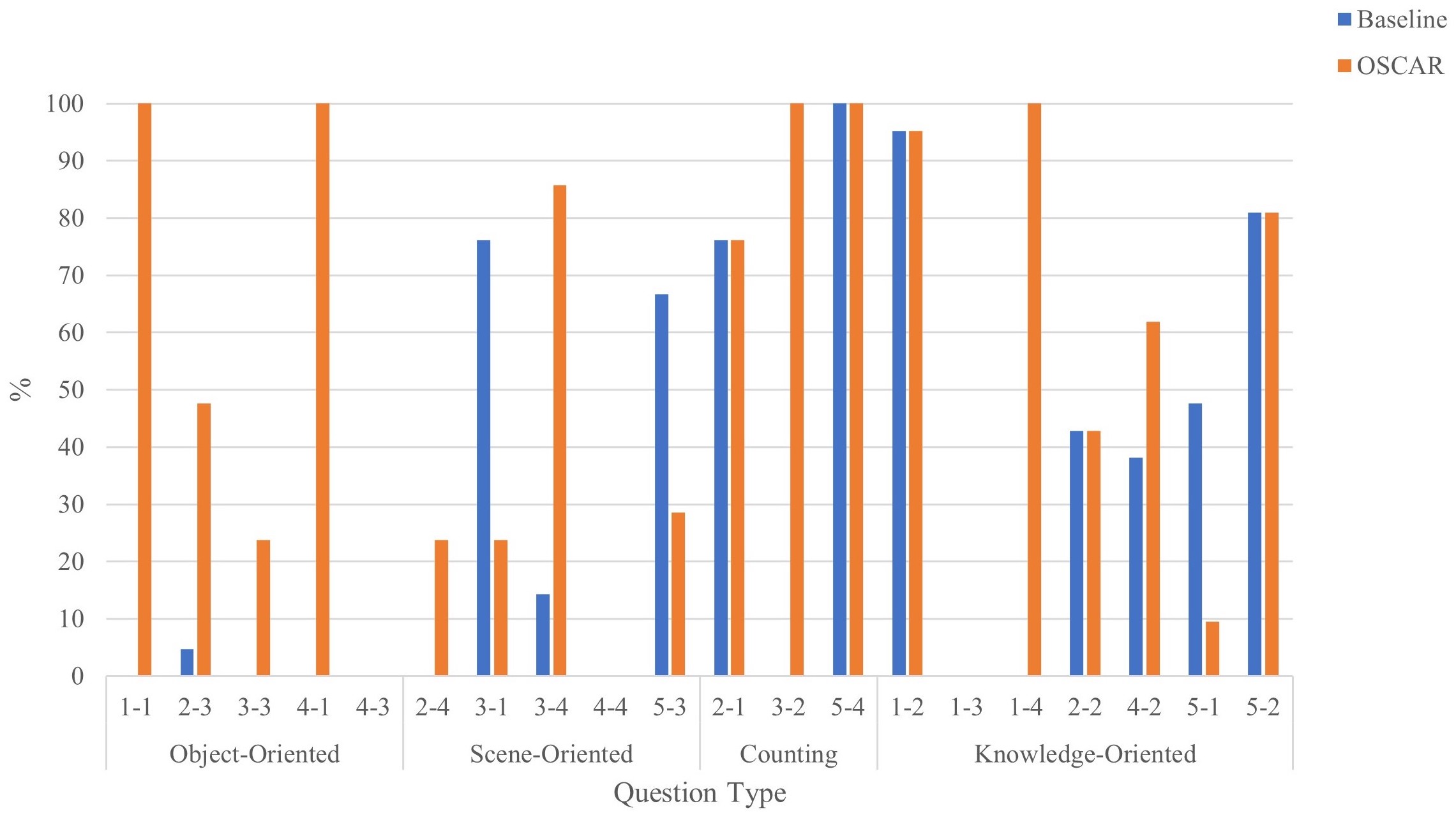}
        \vspace{-.5cm}
        \caption{Grouped by question type.}
        \label{qtypepercent}
    \end{subfigure}
    \begin{subfigure}{\linewidth}
        \includegraphics[width=\textwidth]{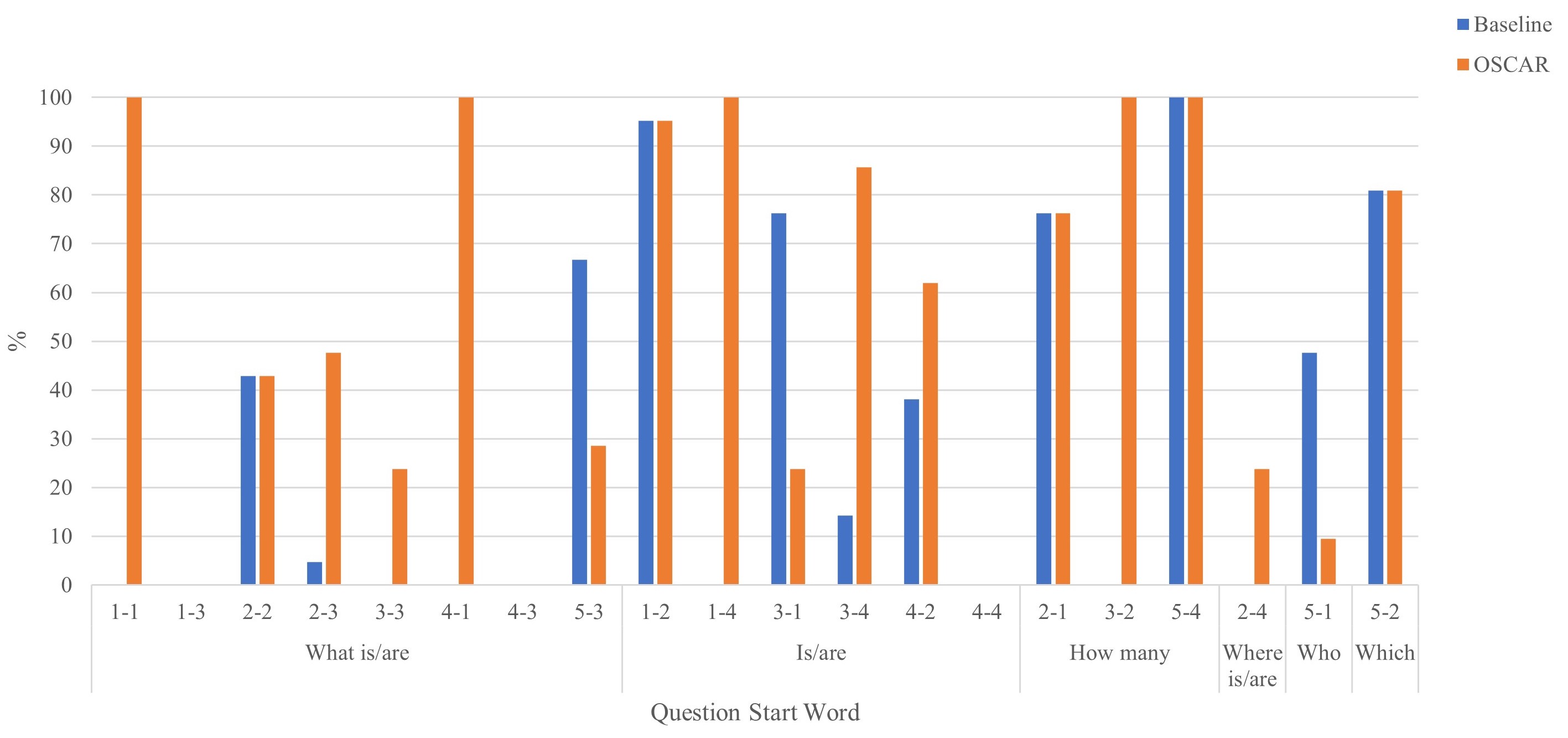}
        \vspace{-.5cm}
        \caption{Grouped by starting words.}
        \label{swordpercent}
    \end{subfigure}
    \caption{Percentage answer alignment between model and human, with three grouping strategies.}
    \label{fig:answer_alignment}
\end{figure}

In this section, we compare human and machine answers to the VQA questions directly in Table.\ref{tab:survey_result}. We first plot the percentages of answers in each question that align with the model prediction in Fig.\ref{imgnumpercent}, Fig.\ref{qtypepercent} and Fig.\ref{swordpercent}. The plots are grouped by the image number, and question types in Table.\ref{tab:qtype} and starting words~\cite{antol2015vqa} that are relevant to answer alignment, as discussed in other papers. At this moment, we consider the answers aligned if the human answers include the exact word in machine predictions. All of these plots only give us a qualitative understanding of the potential trends in the alignment. They assist us to make decisions on further exploring or discarding the factors we once are interested in.

Outputs from both models, as shown in Table.~\ref{tab:survey_result}, indeed present some alignment and capture a moderate amount of keywords in human answers. Generally, OSCAR produces a higher percentage alignment than the baseline model. For different question types, the baseline model does not perform well on the object-oriented questions and also neglects information in scene-oriented and knowledge-oriented questions. In these categories, OSCAR tends to give either close to perfect alignment, or a complete miss. For starting words, since not every question is phrased the same way under the same question type, and vice versa, the distinction in starting words of the questions is relevant to question types as well as the expected form of answers. In Fig.\ref{swordpercent}, most of the questions start with "what is/are". The baseline model outputs are hardly consistent with the human ones in "what" questions and even OSCAR have 0 alignments in 2 of the questions. This may be explained by the observation that humans have a wide variety in their answers to free-formed, particularly object-oriented, questions, or the models are completely off in detecting the object of interest due to their designs.

We also looked into the quality and style of the selected images as shown in Fig.\ref{imgnumpercent}. Image 5 seems to have more alignment than the rest of the images to the baseline. One possible explanation is that the questions we ask in image 5 belong to scene-oriented, counting, and knowledge-oriented categories, but not the object-oriented type which gives the worst performance to the model. Both models miss some questions in Images 1 and 4, while the baseline fails to capture half of the questions in Image 3 especially. We are particularly interested in the questions that the models almost completely lose (less than $10\%$) to the human answers, aiming to tentatively determine what may confuse the models.

We listed the questions in which the baseline has less than $10\%$ alignment to human answers in Table.\ref{base0align}, and OSCAR in Table.\ref{oscar0align}. Consistent with our findings, most of the questions the baseline model losing information belongs to object-oriented ones and begins with "what". OSCAR miss 2 "what" questions out of 4 as well as 2 knowledge-oriented questions. Both models answer incorrectly to Image 1 question 3, Image 4 Question 3, and Image 4 Question 4. In Image 1 Question 3, while people identify the baby as answering or holding the phone, the models predict the baby is brushing teeth or eating ice cream. Notably, these 3 types of postures all require the baby to have a hand near the face. Image 4 has a blurry background composed of multiple colors. The bear does have beige-white and grey components on it, but it is still viewed by humans as a brown bear. We may conjecture that the models look at the images by pixels, and they mainly use their pixel-level understanding and memory from their training set to infer. To get a further observation, we need to look into the model attention.

\begin{table*}[ht!]
\begin{center} 
\vskip 0.12in
\begin{tabular}{lll} 
\hline
Image    &  Question & Question type\\
\hline
1 & (1) What is in the image? & Object\\
1 & (3) What is this baby doing? & Knowledge\\
1 & (4) Is this baby a boy or girl? & Knowledge\\
2 & (3) What food is on the table? & Object\\
2 & (4) Where is the TV? & Scene\\
3 & (2) How many people are there? & Counting\\
3 & (3) What color are the umbrellas on the left? & Object\\
4 & (1) What animal is in the image? & Object\\
4 & (3) What color is the bear’s fur? & Object\\
4 & (4) Is the bear in the river? & Scene\\
\hline
\end{tabular} 
\end{center}
\vspace{-.3cm}
\caption{Questions at the baseline has less than 10\% alignment with human answers.}
\label{base0align} 
\end{table*}

\begin{table}[ht!]
\begin{center} 
\vskip 0.12in
\begin{tabular}{lll} 
\hline
Image    &  Question & Question type\\
\hline
1        &   (3) What is the baby doing? & Knowledge\\
4 & (3) What color is the bear’s fur? & Object\\
4 & (4) Is the bear in the river? & Scene\\
5 & (1) Who is older? & Knowledge\\
\hline
\end{tabular} 
\end{center}
\vspace{-.3cm}
\caption{Questions OSCAR has less than 10\% alignment with human answers.}
\vspace{-.5cm}
\label{oscar0align} 
\end{table}

In some questions, the baseline model gives higher alignment than OSCAR. In most of these cases, OSCAR's answer cannot be conceived as wrong under human perception. For example, in Image 5 Question 1, OSCAR answers "person on the left", while most participants choose to describe the gender of the older person. In image 5 question 3, OSCAR chooses a window rather than a TV, while the window is also next to people in the image. We wonder if using a single standard answer to judge model alignment with humans is appropriate. This may also be a problem in the model training process.

The above results inform us that models with different designs have the possibility to reach a certain level of performance close to humans. Question types, starting words and image varieties may all affect their abilities on VQA tasks. The measurement of model performances may also be influenced by the vagueness in images and questions which lead to large variations in human answers. Carrying on these results, we study the attention map corresponding to the image-question pairs, which further uncovers model and human reasoning similarities and differences. 

\subsection{Attention Map}
Both models addressed in this project use the attention mechanism. Attention map helps us visualize the most important part in the pictures marked by models. The model attention shows its learned representation and understanding of the given textual and visual information as well as other relational features, suggesting the reason behind its prediction. Echoing the model attention map, we also created human attention maps from our survey and observed that the human attention shift on the images before and after seeing the corresponding questions. Meanwhile, we compared the human attention maps to that from our more primitive baseline model to examine human and model thought paths.

From Sub-question 1 in the survey, we first created the attention map of human participants on the image before seeing any of the questions, as in Fig.\ref{fig:global_attention}. We inspected that human put their attention on the most meaningful object~\cite{henderson2017meaning}. Also, a definite, large object on focus~\cite{proulx2010size} may often become the most informative object to humans. Specifically, if such an object is a human or a living entity, people tend to look at the face or especially the eyes. This phenomenon is also reported by many participants in Image 1 Question 1, in which they explicitly mentioned the baby's eyes are attractive and her facial expression is scared or calm. Another theory that may explain our figure is the stimulus motivated attention~\cite{cogpsy}, in which people are able to quickly identify the object with the most special features first~\cite{gelade1980}. Based on this theory, people may also look more at the areas that have more colors, higher contrasts, and brighter light intensity. As we plotted the attention maps from models before viewing the questions for comparison, we notice they have some similarities with those from humans in images with a large object in focus like Image 1 and Image 4. However, the model puts relatively scattered attention on the rest of the images. These attention maps might suggest that humans have prepared some understanding and story-telling for information in the visual materials~\cite{saletta2020role} already before getting any textual guidance, while machines are prone to look into the largest object.

We use Sub-question 3 to extract attention maps and observe shifts in human attention after getting the textual information in Fig.\ref{fig:human_attention}. If the focus areas of humans before given the question have enough information to give the answer, the attention areas do not change; otherwise, they shift. The human may also concentrate their attention on smaller areas based on the text. Interestingly, human attention is not limited to one spot in the image. In Image 3 Question 1, people focus on both the puddle on the ground and the background to determine whether it is raining.

We compared the human attention to our model attention after questions to uncover the model thought process in accordance with humans in Fig.\ref{fig:image_1}, Fig.\ref{fig:image_2}, Fig.\ref{fig:image_3}, Fig.\ref{fig:image_4} and Fig.\ref{fig:image_5}. We can observe differences in their attention maps in multiple questions. To clarify our analysis, we parsed the comparison between the model and the human into 3 situations and raise examples to demonstrate each of them:

\textbf{1. Model attention is correct under human judgment and the answer aligns with the human majority:} This happens only in about 4 out of 20 pairs. In Image 1 Question 2, both the model and the baby focus on the baby's face to check if the baby is not crying. Still, since there is a large portion of attention given to the baby's belly and the model gives attention to the baby's head before the question, we doubt if the model fully understands the meaning of crying to answer the question correctly.

\textbf{2. Model attention is correct, but the answer is not aligned:} Various situations appear for this case. First, the model does not understand the meaning of the object of attention it considers important. In Image 1 Question 1, even though the model focuses on the baby, it still thinks there is a balloon. Similarly, in Image 4 Question 1, it pays attention to the bear but recognizes it as a dog. We doubted this can result that although the model is capable of differentiating objects, catching the focus of the image, and extracting some keywords about the question type, it lacks the ability to associate the extracted information together. Second, the model cannot capture the preferred answer among several equally correct ones. In Image 2 Question 2, the model sees people as well as what they watch, but it answers TV which is the second-highest possible answer.

\textbf{3. Attention and answer both do not align with human judgment:} In this case, the models are completely off in their attention, while the attention is scattered over the entire image, showing the model may not understand what the question asks for.

With the attention maps, we can check if models identify the subject in the picture just as humans do. Models rely heavily on pixel-level visual information and specific keyword extraction on the questions. The design of features and architectures determine the information models are able to process and identify what they see with their memory of the training set. However, it is challenging for models to use commonsense, association, and semantic understanding to correctly infer the expected question answers. Other experiments like ~\citeA{cadene2019murel} show that the model memorizes from the training dataset that bananas are associated with yellow color, without actually understanding its color. Hence, when provided with a green banana during the test, it predicts its color to be yellow. Meanwhile, \citeA{agrawal2016analyzing} shows that the model predicts the car has a license plate, even without looking at the image~\cite{ayyubi2020leveraging}. In addition, their relatively succinct answers differentiate themselves from human's ability to describe their thoughts when they consider a vagueness exists in an image-question pair.

We have obtained some knowledge that humans and models think in different ways from all the results discussed in the previous sections. This is within our expectation since the ultimate goals of the models are not to completely recreate human intelligence and the learning structures, such as the gradient descent optimizations which are not the exact way brains and neurons function. We consider the more significant directions for researchers to further develop their models follow as:

\begin{outline}
    \vspace{-.2cm}
    \1 How can we get more inspiration from human cognition to develop new features and architectures to improve model performances?
    \vspace{-.2cm}
    \1 How can we apply more complicated, multi-discipline questions that are intuitive, and natural for human to technical, engineering methods? 
\end{outline}

Based on these ideas, we attempted to analyze our participants' thought processes in answering the survey questions to gain a comprehensive story of human perception of VQA tasks.

\subsection{Human Thinking Process}
We summarized human perceptions of VQA tasks into 5 stages based on our research, understanding of human cognition, and the previous results. These 5 stages include but are not limited to the 3 perspectives in our hypothesis. They have also been demonstrated in the previous sections with our survey results in the Sub-questions 3 and 4 in Table.\ref{embededq}.

\textbf{1. Object visual recognition:} Humans identify objects using their past experiences. They tend to focus on the most protruding features~\cite{cogpsy} and the most informative portions of common sense ~\cite{henderson2017meaning}. Particularly, if a living entity is in the present, people tend to focus on the living entity's head or eyes. In Image 2 Question 1, many participants reported that they identified the head and postures of people in the image, and then counted by heads. In Image 1 Question 1, a few participants mentioned the baby's eyes. These areas are not identified independently from the rest of the image, but rather comparatively salient. 

\textbf{2. Global attention on the context:} Humans can recognize objects and scenes better if they are put into a background or context that is commonly seen with the object. They look into the image as a whole scene rather than split it into independent patches~\cite{biederman1973searching}. In Image 3 Question 3, even though there is no explicit raindrop in the image, participants justified their answers by combining many factors such as umbrellas, wet ground, and puddles to collaboratively answer yes to the question.

\textbf{3. Domain knowledge and experience:} Humans use commonsense and knowledge abreast with their visual stimulation to infer, justify and polish their answers. This can be done sometimes even when the seen object is not typical in their past experiences. For example, in Image 1 Question 2, 10 participants mentioned no tears, and 10 mentioned facial expressions such as calmness among 18 responses to justify their answers of "not crying". 

\textbf{4. Narrative generation and creativity:} Humans automatically generate a full scenario or a personal narrative after viewing the image by inference and knowledge. In Image 1 Question 1, some participants come up with a full description of what the baby is doing in the image. Some mention the baby is a girl because she is wearing pink. She looks helpless and scared and is probably calling a relative.

\textbf{5. Data requirement:} Humans regularly do not require a massive amount of data to learn and use their learned knowledge. Their past learning accumulates daily and applies to future questions actively. 

Meanwhile, these human-specific abilities generate contemplation and variations during the reasoning process that are also worth considering in model design and training. We asked our participants if they found the questions to require more time than immediate in the sub-questions 5 and 6 in Table.\ref{embededq}. We raised questions for which over 1/3 of the participants select level 3 to 5. We noticed 3 situations that confused our participants, overlapping with our discussion at the beginning of the survey results. First, participants point out some questions are phrased too broad and vague to make an exact and unique answer. Instead, they may create a narrative for themselves when they encounter such vagueness. Second, some image-question pairs do not have enough information for people to tie a unified inference. For example, in Image 4 Question 2, almost half of the responses mention they can only guess the bear is roaring, angry, yawning, or sleepy but not hungry. Third, participants reported the subject asked by the question is too detailed or trivial to be quickly found in the image.

In summary, in the survey result section, we extracted the human answers to the representative image-questions pair and find models may reach a moderate alignment with them. We then looked into the attention map of human and model to inform if they have a similar thought process, but found them not often consistent to each other. Humans have more capabilities in using the aforementioned 5 stages of the human thinking process to complete the VQA task. Our result is consistent with our hypothesis in the human thinking process and models.

\section{Conclusion}
We conducted a comparative analysis of the VQA models and human question answering ability in three perspectives: (1) model architecture inspired by the human thought process, (2) visual and textual information understanding, (3) prediction results and reasonings. We developed a baseline model that uses ResNet and LSTM for embedding image and question text with an attention module that has potential counterparts in human perception, meanwhile employing a SOTA transformer model OSCAR for a comparison. We also designed a survey with 20 image-question pairs, each with sub-questions to track the reasoning progress of participants. The model outputs are compared to human results in three steps (1) direct comparison between human and model answers, (2) human and model attention maps and (3) rationales in the decision-making process. The results demonstrate that the VQA model may perform better on recognition-level image understanding, but struggles with complex inferences which result in diverse answers in humans. We argue that as the field has made significant progress on recognition-level building blocks such as object detection, pose estimation, and segmentation, now is the right time to tackle cognition-level reasoning at scale \cite{ayyubi2020generating}.   

From a cognitive science perspective, future works can incorporate the human ways of thinking into the model features and architectures. Several approaches have already been studied, such as zero-shot learning~\cite{teney2016zero}, which promoted the model's ability to answer questions that are out of the scope of training data, simulating humans’ reflective recognition abilities. The Visual Commonsense Reasoning dataset added a justification step after the VQA image-question answering to provide rationales for the predicted answer~\cite{zellers2019recognition}, reproducing the human thought process. \citeA{qiao2018exploring} proposed a Human Attention Network (HAN) to generate human-like attention maps, buttressing that the human-like model supervision would benefit VQA tasks. Also, with the recent development of a large-scale knowledge base, a more fine-grained semantic space can be used for more accurate object detection and textual semantic understanding. Overall, the future success of VQA lies in artificial cognitive abilities.

\bibliographystyle{apacite}
\setlength{\bibleftmargin}{.125in}
\setlength{\bibindent}{-\bibleftmargin}
\bibliography{CogSci_Template}

\onecolumn
\renewcommand\thefigure{A.\arabic{figure}}
\setcounter{figure}{0}
\renewcommand\thetable{A.\arabic{table}}
\setcounter{table}{0}

\section*{Appendix}
\bigskip

\subsection{Human and Baseline Model Attention Visualization}

\begin{figure}[htbp]
    \centering
    \includegraphics[width=.7\textwidth]{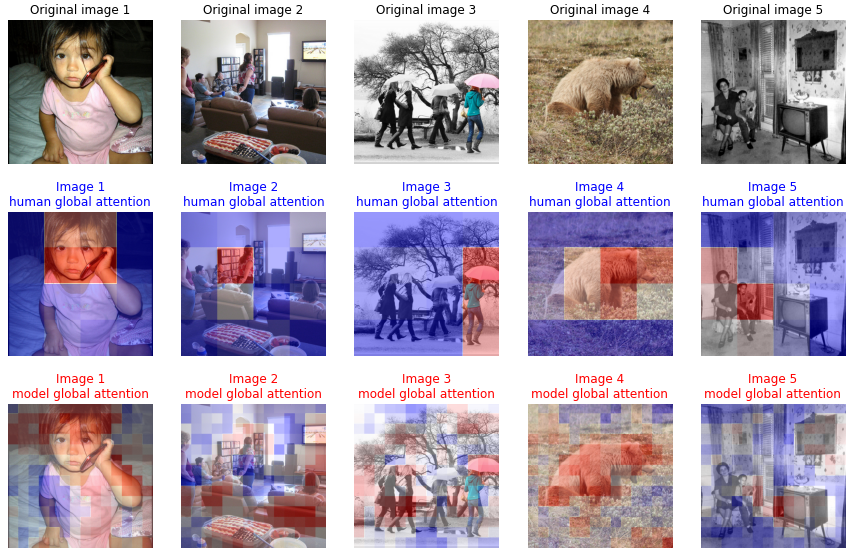}
    \vspace{-.3cm}
    \caption{Global Attention with Original Image.}
    \label{fig:global_attention}
\end{figure}

\begin{figure}[htbp]
    \centering
    \includegraphics[width=.7\textwidth]{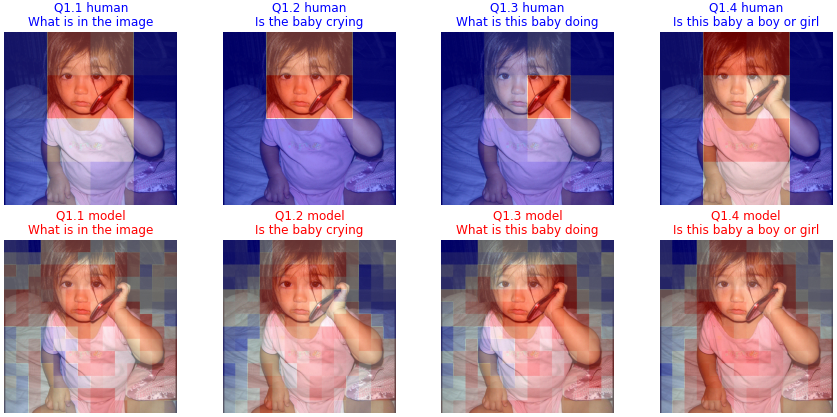}
    \vspace{-.3cm}
    \caption{Human vs. model attention on Image 1.}
    \label{fig:image_1}
\end{figure}

\begin{figure}[htbp]
    \centering
    \includegraphics[width=.7\textwidth]{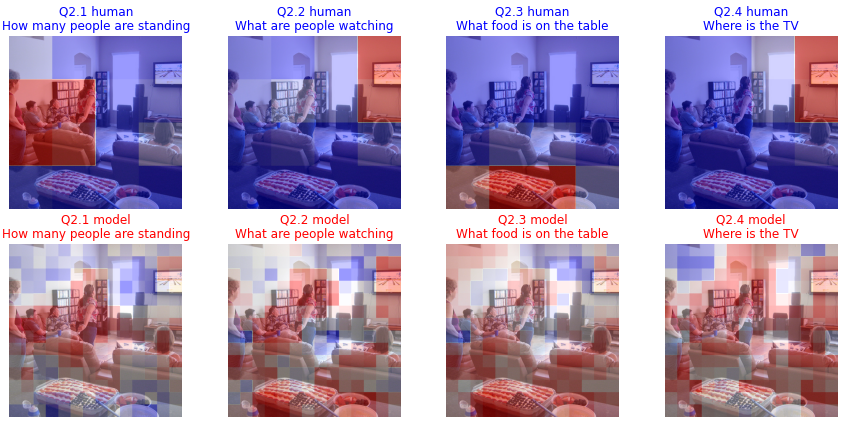}
    \vspace{-.3cm}
    \caption{Human vs. model attention on Image 2.}
    \label{fig:image_2}
\end{figure}

\begin{figure}[htbp]
    \centering
    \includegraphics[width=.7\textwidth]{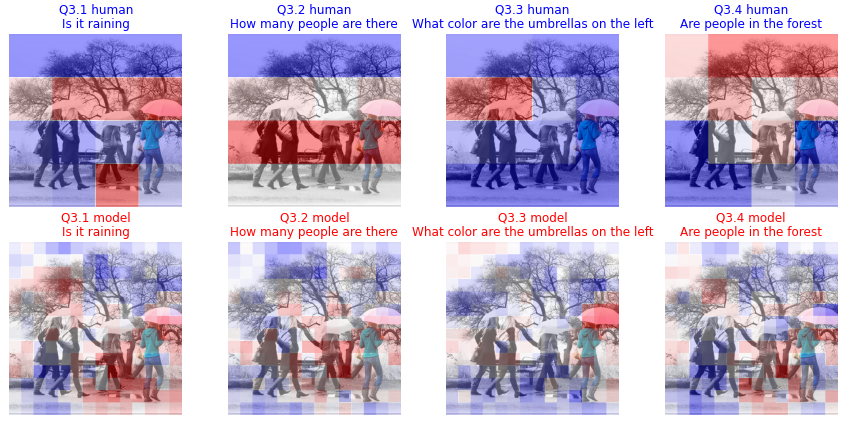}
    \vspace{-.3cm}
    \caption{Human vs. model attention on Image 3.}
    \label{fig:image_3}
\end{figure}

\begin{figure}[htbp]
    \centering
    \includegraphics[width=.7\textwidth]{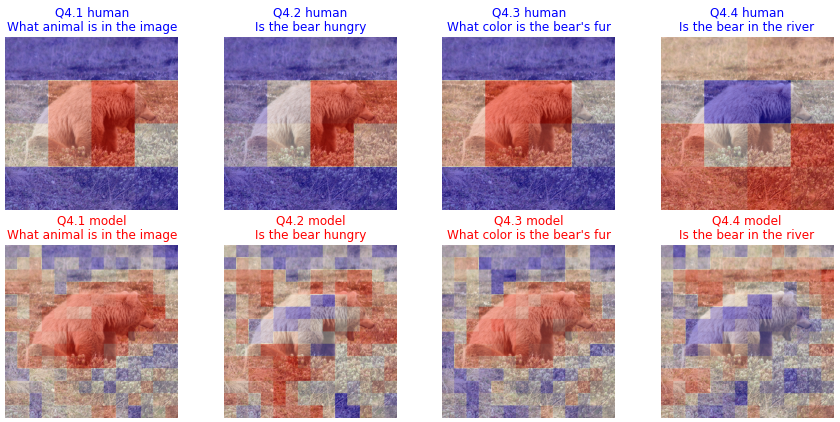}
    \vspace{-.3cm}
    \caption{Human vs. model attention on Image 4.}
    \label{fig:image_4}
\end{figure}

\begin{figure}[htbp]
    \centering
    \includegraphics[width=.7\textwidth]{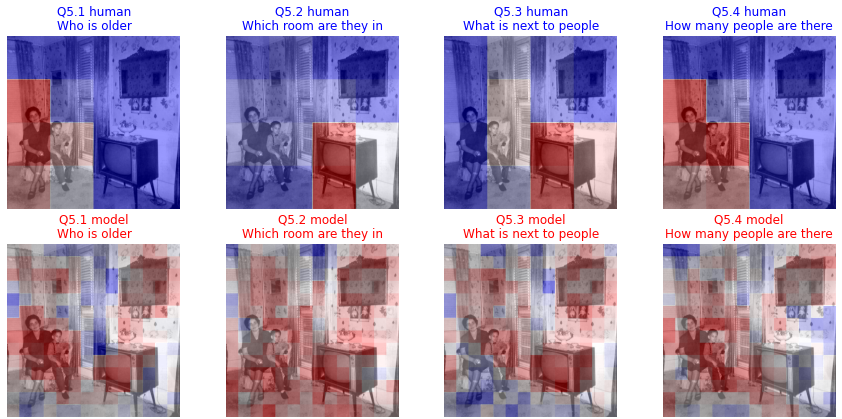}
    \vspace{-.3cm}
    \caption{Human vs. model attention on Image 5.}
    \label{fig:image_5}
\end{figure}

\begin{figure}[htbp]
    \centering
    \includegraphics[width=.8\textwidth]{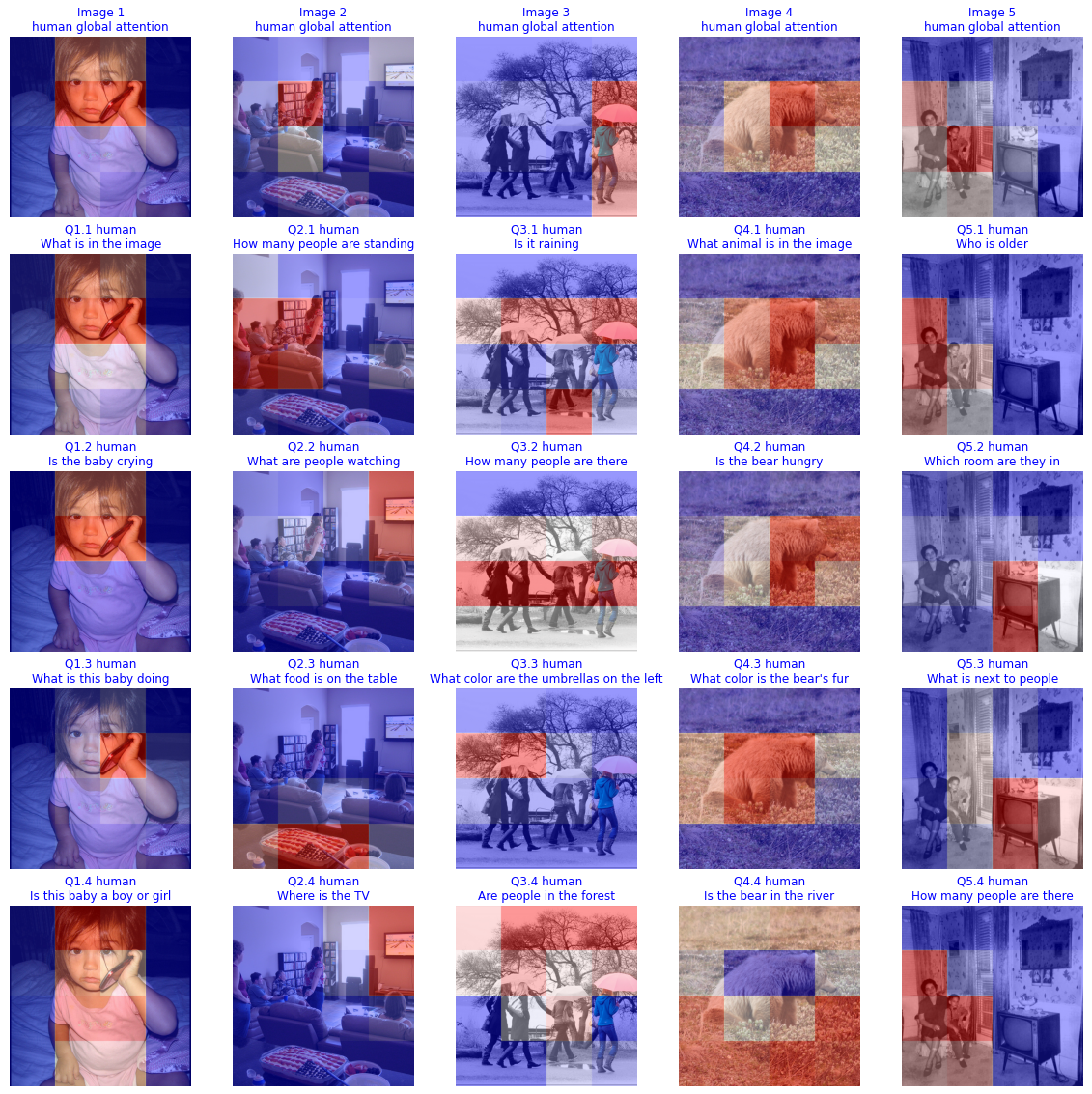}
    \vspace{-.3cm}
    \caption{Human attention comparison for all images.}
    \label{fig:human_attention}
\end{figure}

\newpage
\subsection{Survey Results and Model Outputs}

\begin{longtable}{|p{.05\textwidth}|p{.2\textwidth}|p{.2\textwidth}|p{.12\textwidth}|p{.13\textwidth}|p{.13\textwidth}|}
    \hline
    \textbf{Image} & \textbf{Question} & \textbf{Key Words and Phrases in Human Answer} & \textbf{Percentage of Appearances} & \textbf{Baseline Model Answer} & \textbf{OSCAR Model Answer} \\
    \hline
    \endhead
    \multirowcell{10}[0pt][l]{1} & \multirowcell{2}[0pt][l]{(1) What is in the image?} & Girl & 100.00 & \multirowcell{2}[0pt][l]{Balloon}& \multirowcell{2}[0pt][l]{Girl}\\
    \cline{3-4}
    & & Girl, phone & 61.90 &  & \\
    \cline{3-4}
    & & Girl, in the dark & 23.80 &  & \\
    \cline{3-4}
    & & Girl, calling & 4.76 &  & \\
    \cline{2-6}
    & \multirowcell{2}[0pt][l]{(2) Is this baby crying?} & No & 95.24 & \multirowcell{2}[0pt][l]{No}& \multirowcell{2}[0pt][l]{No}\\
    \cline{3-4}
    & & Yes & 4.76 & & \\
    \cline{2-6}
    & \multirowcell{2}[-6pt][l]{(3) What is this baby \\doing?} & Talking/Answering phone & 76.19 & \multirowcell{2}[0pt][l]{Brushing teeth}& \multirowcell{2}[-6pt][l]{Eating \\ice-cream}\\
    \cline{3-4}
    & & Yes & 4.76 & & \\
    \cline{2-6}
    & \multirowcell{2}[0pt][l]{(4) Is this baby a boy \\or a girl?} & Girl & 100.00 & \multirowcell{2}[0pt][l]{Boy}& \multirowcell{2}[0pt][l]{Girl}\\
    \cline{3-4}
    & & Boy & 0.00 &  & \\
    \cline{1-6}
    
    \multirowcell{20}[0pt][l]{2} & \multirowcell{3}[0pt][l]{(1) How many people \\are standing?} & 2 & 76.19 & \multirowcell{3}[0pt][l]{2}& \multirowcell{3}[0pt][l]{2}\\
    \cline{3-4}
    & & 5 & 19.05 &  & \\
    \cline{3-4}
    & & 5 guests + 1 chef & 4.76 &  & \\
    \cline{2-6}
     & \multirowcell{5}[0pt][l]{(2) What are people \\ watching?} &Bowling &47.62 & \multirowcell{5}[0pt][l]{TV}& \multirowcell{5}[0pt][l]{Television}\\
     \cline{3-4}
    & & TV & 42.86 &  & \\
    \cline{3-4}
    & & Five & 9.52 &  & \\
    \cline{3-4}
    & & Bowling on TV & 4.76&  & \\
    \cline{3-4}
    & & Parade & 4.76 &  & \\
    \cline{2-6}
    & \multirowcell{8}[0pt][l]{(3) What food is on \\the table?} & Fruit & 66.66& \multirowcell{8}[0pt][l]{Pizza}& \multirowcell{8}[0pt][l]{Pie}\\
    \cline{3-4}
    & & Pie & 47.63 &  & \\
    \cline{3-4}
    & & Cake & 28.57 &  & \\
    \cline{3-4}
    & & Dessert & 9.52 &  & \\
    \cline{3-4}
    & & Pasta & 4.76 &  & \\
    \cline{3-4}
    & & Pizza & 4.76 &  & \\
    \cline{3-4}
    & & Fruit cream box & 4.76 &  & \\
    \cline{3-4}
    & & Not sure& 4.76 &  & \\
    \cline{2-6}
    & \multirowcell{4}[0pt][l]{(4) Where is the TV?} & Wall & 90.48 & \multirowcell{4}[0pt][l]{Top left}& \multirowcell{4}[0pt][l]{On the right}\\
    \cline{3-4}
    & & Up right corner & 23.81 &  & \\
    \cline{3-4}
    & & On the wall in the living room & 4.76 &  & \\
    \cline{3-4}
    & & On the wall beside window & 4.76 &  & \\
    \cline{1-6}
    
    \multirowcell{10}[0pt][l]{3}& \multirowcell{2}[0pt][l]{(1) Is it raining?} & Yes & 76.19& \multirowcell{2}[0pt][l]{Yes}& \multirowcell{2}[0pt][l]{No}\\
    \cline{3-4}
    & & No & 23.81 &  & \\
    \cline{2-6}
    & (2) How many people are there? & 4& 100.00& 2 & 4\\
    \cline{2-6}
    & \multirowcell{5}[0pt][l]{(3) What color are the \\umbrellas on the left?} & White & 42.86 & \multirowcell{5}[0pt][l]{Red}& \multirowcell{5}[0pt][l]{Grey}\\
    \cline{3-4}
    & & Grey& 23.81 &  & \\
    \cline{3-4}
    & & Not sure & 14.29 &  & \\
    \cline{3-4}
    & & Light color & 9.52 &  & \\
    \cline{3-4}
    & & Pink & 4.76 &  & \\
    \cline{2-6}
    & \multirowcell{2}[0pt][l]{(4) Are people in \\the forest?} &No &85.71 & \multirowcell{2}[0pt][l]{Yes}& \multirowcell{2}[0pt][l]{No}\\
    \cline{3-4}
    & & Yes & 14.29 &  & \\
    \cline{1-6}
    
    \multirowcell{10}[22pt][l]{4}& \multirowcell{3}[0pt][l]{(1) What animal is in \\the image?} & Bear & 100.00& \multirowcell{3}[0pt][l]{Dog}& \multirowcell{3}[0pt][l]{Bear}\\
    \cline{3-4}
    & & A bear & 9.52 &  & \\
    \cline{3-4}
    & & Brown bear & 9.52 &  & \\
    \cline{2-6}
    & \multirowcell{2}[0pt][l]{(2) Is the bear hungry} & No & 61.90 & \multirowcell{2}[0pt][l]{Yes}& \multirowcell{2}[0pt][l]{No}\\
    \cline{3-4}
    & & Yes & 38.10 & & \\
    \cline{2-6}
    \pagebreak
    & \multirowcell{4}[0pt][l]{(3) What color is the \\bear's fur?} &Brown &90.48 & \multirowcell{4}[0pt][l]{White}& \multirowcell{4}[0pt][l]{Grey}\\
    \cline{3-4}
    & & Light brown & 28.57 &  & \\
    \cline{3-4}
    & & Beige & 9.52 &  & \\
    \cline{3-4}
    & & Dark beige & 4.76 &  & \\
    \cline{2-6}
    & (4) Is the bear in the river? & No & 100.00 & Yes & Yes\\
    \cline{1-6}
    
    \multirowcell{17}[0pt][l]{5}& \multirowcell{7}[0pt][l]{(1) Who is older?} & Woman & 47.62 & \multirowcell{7}[0pt][l]{Woman on left}& \multirowcell{7}[0pt][l]{Person on the \\left}\\
    \cline{3-4}
    & & Left woman & 23.81 &  & \\
    \cline{3-4}
    & & Lady & 23.81 &  & \\
    \cline{3-4}
    & & Female & 9.52 &  & \\
    \cline{3-4}
    & & Person on the left & 9.52 &  & \\
    \cline{3-4}
    & & Girl & 4.76 &  & \\
    \cline{3-4}
    & & Mother & 4.76 &  & \\
    \cline{2-6}
    & \multirowcell{4}[0pt][l]{(2) Which room are \\they in?} & Living room & 80.95& \multirowcell{4}[0pt][l]{Living room}& \multirowcell{4}[0pt][l]{Living room}\\
    \cline{3-4}
    & & Bedroom & 4.76 &  & \\
    \cline{3-4}
    & & Family room & 4.76 &  & \\
    \cline{3-4}
    & & Den & 4.76 &  & \\
    \cline{2-6}
    & \multirowcell{5}[0pt][l]{(3) What is next to \\people?} & TV & 66.66 & \multirowcell{5}[0pt][l]{TV}& \multirowcell{5}[0pt][l]{Window}\\
    \cline{3-4}
    & & Window & 28.57 &  & \\
    \cline{3-4}
    & & TV or curtain & 4.76 &  & \\
    \cline{3-4}
    & & Door & 4.76 &  & \\
    \cline{3-4}
    & & Maybe TV & 4.76 &  & \\
    \cline{2-6}
    & (4) How many people are there? & 2 & 100.00 & 2 & 2\\
    \hline
    \caption{Results from survey and outputs from Baseline and OSCAR models.}
    \label{tab:survey_result}
\end{longtable}

\end{document}